\documentclass[10pt,journal,onecolumn]{IEEEtran}
\usepackage{amsmath,amssymb,graphicx}
\usepackage{subfigure}
\usepackage{multirow}

\begin{document}

\title{Channel Max Pooling Layer\\ for Fine-Grained Vehicle Classification}

%


\author{Zhanyu Ma, Dongliang Chang, and Xiaoxu Li}


\maketitle

\begin{abstract}
Deep convolutional networks have recently shown excellent performance on Fine-Grained Vehicle Classification. Based on these existing works, we consider that the back-probation algorithm does not focus on extracting less discriminative feature as much as possible, but focus on that the loss function equals zero. Intuitively, if we can learn less discriminative features, and these features still could fit the training data well, the generalization ability of neural network could be improved. Therefore, we propose a new layer which is placed between fully connected layers and convolutional layers, called as Chanel Max Pooling. The proposed layer groups the features map first and then compress each group into a new feature map by computing maximum of pixels with same positions in the group of feature maps. Meanwhile, the proposed layer has an advantage that it could help neural network reduce massive parameters. Experimental results on two fine-grained vehicle datasets, the Stanford Cars-196 dataset and the Comp Cars dataset, demonstrate that the proposed layer could improve classification accuracies of deep neural networks on fine-grained vehicle classification in the situation that a massive of parameters are reduced. Moreover, it has a competitive performance with the-state-of-art performance on the two datasets.
\end{abstract}


\IEEEpeerreviewmaketitle

\section{Introduction}

Even though computer vision has achieved big improvements in all kinds of tasks, it still exists a big challenge on Fine-grained visual classification (FGVC). The difficulty of FGVC, which aims to distinguish subordinate visual categories, is mainly due to big intra-class similarity. So far, FGVC has been paid much attention on classification of birds, dogs, flowers, vehicle, and so on. In this paper, we focus on the fine-grained vehicle classification.

With the successful application of deep convolutional neural network in all kinds of task of computer vision, much works have taken advantage of CNN-features for fine-grained classification. Krause et al. \cite{krause2015fine} proposed a method for fine-grained recognition that uses no part annotations and generates parts which can be detected in new images and learn which parts are useful for recognition. Simon et al. \cite{simon2015neural} presented an approach that is able to learn part models in a completely unsupervised manner, without part annotations and even without given bounding boxes during learning. Xiao et al. \cite{xiao2015application} presented a fine-grained classification pipeline combining bottom-up and two top-down attentions, and the weak supervision constraint makes our work easier to generalize. Zhang et al. \cite{zhang2016picking} propose a framework that incorporates deep convolutional filters for both part detection and description for fine-grained recognition, which is free of any object annotation at both training and testing stages.

Besides, one category of localization-classification subnetworks, mainly utilize part locations or segmentation masks provided by a localization network to improve the performance of classification network \cite{wang2018learning}. Zhang et al.\cite{zhang2014part} proposed a system for joint object detection and part localization capable of state-of-the-art fine-grained object recognition. Our method learns detectors and part models and enforces learned geometric constraints between parts and with the object frame. Fu et al. \cite{fu2017look} propose a recurrent attention convolutional neural network for fine-grained recognition, which recursively learns discriminative region attention and region-based feature representation at multiple scales. The proposed network does not need bounding box/part annotations for training and can be trained end-to-end. \cite{huang2016part} propose a novel Part-Stacked CNN architecture that explicitly explains the fine-grained recognition process by modeling subtle differences from object parts. Jaderberg et al. \cite{jaderberg2015spatial} introduced the spatial transformer, which can be dropped into a network and perform explicit spatial transformations of features. and performed well on fine-grained classification. \cite{lam2017fine} proposed a search-based architecture where the search space is defined on a convolutional feature map of CNN a search-based framework with deep architectures for fine-grained recognition that achieves competitive results. Lin et al. \cite{lin2015deep} proposed a fine-grained recognition system that incorporates part localization, alignment, and classification in one deep neural network. Zhang et al. \cite{zhang2016spda} proposed a new CNN architecture that integrates semantic part detection and abstraction  for fine-grained classification.

Another category of end-to-end feature encoding encodes higher order statistics of convolutional feature maps to enhance CNN mid-level representation. Lin et al. \cite{lin2015bilinear} propose bilinear models which can model local pairwise feature interactions in a translationally invariant manner which is particularly useful for fine-grained categorization. Gao et al. \cite{gao2016compact} propose two compact bilinear representations with the same discriminative power as the full bilinear representation but with only a few thousand dimensions. Kong et al. \cite{kong2017low} presented an approach for training a very compact low-rank classification model that is able to leverage bilinear feature pooling for fine-grained classification. Cui et al. \cite{cui2017kernel} introduced a novel deep kernel pooling method as a high-order representation for visual recognition. Cai et al. \cite{cai2017higher} is proposed a polynomial kernel based predictor to capture higher-order statistics of convolutional activations for modeling part interaction, which can be plugged in conventional CNNs.

Unlike these existing works, we consider the problem of FGVC from that the back-propagation algorithm simply focuses on that the loss function equals zero, and does not focus on extracting more general feature as much as possible. As a result, the features that the network learned are redundant. Intuitively, if a network could learn more reduced or compressed feature and it still fit training data well, generalization ability could be improved. In addition, we noticed the layer connected convolutional layers and fully connected layer contain massive neurons and connected massive weight parameters. Therefore, we have an idea of adding a new layer that compressed or deducted features while keeping the discrimination capability.

Following the idea, we reduce features by compressing multiple feature maps into a feature map, and proposed channel max pooling layer for deep neural networks with the application in fine-grained vehicle classification. It has two advantages: one is to reduce massive parameter of neural network and make network become simple. Another one is to compress the features and to improve generalization ability of neural network. Experimental results show that on the Stanford Cars-196 dataset and the CompCars dataset, a network employed channel max pooling layer could obtain better performance than the same  network without CMP Layer, and is able to achieve competitive performance with the state-of-the-art performance on the two datasets.

\section{Channel Max Pooling Layer}
In this section, we introduce the proposed channel max pooling layer for convolutional neural networks (CNNs) in fine-grained vehicle classification and name the architecture as CMP-CNN.
\subsection{Channel Max Pooling Layer}
A CMP layer conducts channel-wise max pooling among the corresponding positions of the contiguous feature maps.
Considered as a CNN architecture, a feature extraction function $f_1$ is a mapping function as $f_1:R^{3\times W\times H}\rightarrow R^{C\times M\times N}$, where $W$ and $H$ are the width and the height of the input RGB images $\mathbb{I}$, respectively, and $C$, $M$, and $N$ are the channel number, the width, and the height of the extracted feature maps $\mathbb{F}$. Then, a CMP layer $f_{\mathcal{CMP}}$ can be considered as a mapping $f_{\mathcal{CMP}}:R^{C\times M\times N}\rightarrow R^{\left \lceil\frac{C}{r}\right \rceil\times M\times N}$ , where $\left\lceil\cdot\right\rceil$ means ceiling function, and obtain the compressed feature maps $\mathbb{C}$, where $r$ is the compression factor which will be defined in Section \ref{ssec:cmp}.


\subsection{One CNN with Channel Max Pooling Layer}\label{ssec:cmpcnn}
A CMP-CNN for fine-grained vehicle classification contains three parts, \emph{i.e.}, the feature extraction function $f_1$, the CMP layer $f_{\mathcal{CMP}}$, and the classification function $f_2$. Here, $f_1$ can be considered as not only existing CNN architectures, but also other feature extraction models. Meanwhile, the $f_2$ is usually applied the fully-connected layers. In this paper, we only discuss  $f_1$ as convolutional layers.



\subsection{Discussion}\label{ssec:cmp}

Two hyper parameters are required for the CMP layer which are compression factor $r$ and stride $s$. Here, $r$ is used to determine the compression rate from the input extracted feature maps $\mathbb{F}\in R^{C\times M\times N}$ to the output compressed feature maps $\mathbb{C}\in R^{\left \lceil\frac{C}{r}\right \rceil\times M\times N}$. Meanwhile, $s$ is used to control the stride of max pooling operator in the CMP layer. Moreover, another parameter called kernel size $k$ of max pooling operator is calculated by $r$ and $s$ as
\begin{equation}
k=C-s\cdot\left(\left\lceil\frac{C}{r}\right\rceil-1\right),\label{eq:calk}
\end{equation}
\noindent where $C$ and $s$ are integers larger than $1$, and $r$ is a real number larger than 1.








Deep convolutional neural networks (DCNNs) are trained using the backpropagation algorithm. In order to end-to-end train the DCNNs with CMP, the CMP layer should be derivable. Similar with max pooling layer, the error signals in the CMP layer are only propagated to the position at the maximum value as described in \cite{scherer2010evaluation}.

\section{Experimental Results}
\subsection{Fine-grained Image Datasets}
To evaluate the proposed CMP Layer
In this section, we comprehensively evaluate the cmp method on the Cars-196 \cite{krause20133d} and CompCars \cite{yang2015large}, which are widely-used to evaluate fine-grained Vehicle Classification. These two datasets cover a large number of vehicles of various models in different scenes. The detailed statistics of category numbers and dataset splits are summarized in Table I. Previous studies have shown that using pre-trained DCNNs would lead to good performance on general object classification tasks. Therefore, in our experiments, we apply the proposed CMP layer to those pre-trained DCNNs to further improve the performance of the fine-grained Vehicle Classification.

\begin{table}[htbp]
	\centering
	\caption{The statistics of fine-grained datasets in this paper.}
	\begin{tabular}{l|c|c|c}
		\hline
		Datasets & \multicolumn{1}{l|}{\#Category} & \multicolumn{1}{l|}{\#Training} & \multicolumn{1}{l}{\#Testing} \\
		\hline
		Cars-196 & 196   & 8,144 & 8,041 \\
		CompCars & 431   &  36,456 & 15,627  \\
		\hline
	\end{tabular}%
	\label{tab:addlabel}%
\end{table}%

\subsection{Network Architectures and Implementation Details}
We consider three  widely used pre-trained DCNNs: DenseNet161,
VGG16, and ResNet152 as baselines to evaluate the performance improvement using the proposed CMP layer. In addition to the ResNets152 with default structures, we introduce different variants of ResNets as the improved baselines, which will be described in below.

1) Pre-trained DenseNet161(DenseNet161): DenseNet161  consists of two parts, features and classifier. We replaced the classifier layer trained on ImageNet dataset with a randomly initialized fully connected  layer with a batch normalization (BN) layer, a drop-out layer (dropout probability of $0.5$), and a ELU layer as the activation function, which hidden nums is 256 for Cars-196 and 512 for CompCars, and a randomly initialized
$P$-way softmax layer, where $P$ is the number of classes in the specific dataset.

2) DenseNet161 with CMP (DenseNet161-CMP): As described in Section III, we apply the CMP layer in the DenseNet161 by inserting the CMP layer between the last max-pooling layer and the first fully connected(FC) layer, the dimension of the pooled features is changed from 7 $\times$ 7  $\times$ 2208 to 7  $\times$ 7  $\times$ $\frac{C}{r}$. Meanwhile, all other components remain unchanged.

3) Pre-trained VGG16(VGG16): Similar with DenseNet161, We replaced the classifier layer trained on ImageNet dataset with a randomly initialized fully connected  layer with a batch normalization (BN) layer, a drop-out layer, and a ELU layer as the activation function, which hidden nums is 256 for Cars-196 and 512 for CompCars, and a randomly initialized $P$-way softmax layer, where $P$ is the number of classes in the specific dataset.

4) VGG16 with CMP (VGG16-CMP): Similar with DenseNet161-CMP,
we apply the CMP layer in the VGG16 by inserting the CMP layer between the last max-pooling layer and the first fully connected(FC) layer, the dimension of the pooled features is changed from 7  $\times$ 7  $\times$ 512 to 7  $\times$ 7  $\times$ $\frac{C}{r}$. All other components remain unchanged.

5) ResNet152 without the global average pooling (ResNet152-WoGAP): Instead of using the local max-pooling layer, ResNet152 use the global average pooling (GAP) layer as the last pooling layer. However, the pooled results of the GAP lose all spatial information about the objects and may be too spatially coarse (obtaining feature maps size as 1  $\times$ 1  $\times$ 2048 via a 7  $\times$ 7 average pooling kernel) to capture fine-grained information  of the input image. To remedy this problem, we remove the last GAP layer same as \cite{hu2017dee}. Since the dimension of the pooled features is changed from 1  $\times$ 1  $\times$ 2048 to 7  $\times$ 7  $\times$ 2048, we replaced the classifier layer with a randomly initialized fully connected  layer with a batch normalization (BN) layer, a drop-out layer, and a ELU layer as the activation function, which hidden nums is 256 for Cars-196 and 512 for CompCars, and a randomly initialized $P$-way softmax layer, where $P$ is the number of classes in the specific dataset. Meanwhile, other parts of the model are unchanged. In addition, we refer to ResNet152 with GAP as ResNet152.

6) ResNet152 with CMP (ResNet152-CMP):  We directly apply the CMP layer in the ResNet152-WoGAP between the last convolutional layer  and the first FC layer. The CMP layer change the dimension of pooled features from 1 $\times$ 1 $\times$ 2048 to 7 $\times$ 7 $\times$ $\frac{C}{r}$ where K is the number of compression factor.All other components remain unchanged.

\begin{table*}[htbp]
	\centering
	\caption{COMPARISON OF CLASSIFICATION RESULTS OF VARIOUS DCNNs WITH DIFFERENT PARAMETER SETTINGS ON THE STANFORD CARS-196 DATASET. $``CMP''$ INDICATES THE CHANNEL MAX POOLING.}
	\begin{tabular}{c|c|c|c|c|c|c}
		\hline
		\multicolumn{1}{l|}{compression} & \multicolumn{2}{c|}{DenseNet161-CMP} & \multicolumn{2}{c|}{VGG16-CMP} & \multicolumn{2}{c}{ResNet152-CMP} \\
		\cline{2-7}    rate ($r$)& acc.  & params. & acc.  & params. & acc.  & params. \\
		\hline
		2     & 0.9317 & 54.4M & \textbf{0.9194} & 21.3M & 0.9292 & 84.0M \\
		4     & 0.9297 & 40.5M & 0.9190 & 18.0M   & 0.9292 & 71.1M \\
		8     & 0.9336 & 33.5M & 0.9158 & 16.4M & 0.9285 & 64.7M \\
		16    & \textbf{0.9371} & 30.0M   & 0.9071 & 15.6M & \textbf{0.9296} & 61.5M \\
		32    & 0.9357 & 28.3M & 0.9026 & 15.2M & 0.9263 & 59.9M \\
		\hline
		\hline
		& \multicolumn{2}{c|}{DenseNet161} & \multicolumn{2}{c|}{VGG16} & \multicolumn{2}{c}{ResNet152-WoGAP} \\
		\cline{2-7}    -     & acc.  & params. & acc.  & params. & acc.  & params. \\
		\cline{2-7}          & 0.9292 & 82.2M & 0.9184 & 27.7M & 0.9249 & 109.8M \\
		\hline
	\end{tabular}%
	\label{tab:addlabel}%
\end{table*}%

To adapt the aforementioned DCNNs and baseline DCNNs in the fine-grained vehicle classification, we fine-tune these DCNNs on each car dataset individually, respectively.
We resized the image dimension to 224 $\times$ 224 before inputing into the DCNNs. Then a 224 $\times$ 224 (with padding of 4) crop was randomly sampled from the image or its horizontal flip, with the per-pixel mean subtracted.  For training set, we used the stochastic gradient descent (SGD) with a mini-batch size of 32. Meanwhile, learning rate of FC layers was initialized by $0.1$ and reduced following cosine annealing schedule \cite{loshchilov2016sgd} when training, and learning rate of convolutional layers maintained $\frac{1}{10}$ of FC layers. For all the aforementioned models, we trained for up to $300$ epochs, and used a weight decay of $0.0005$ and a momentum of $0.9$.



\begin{table}[htbp]
	\centering
	\caption{comparison of classification results on the
		cars-196 dataset with bounding box annotations.}
	\begin{tabular}{c|c}
		\hline
		Methods & \multicolumn{1}{l}{ Acc. on Model } \\
		\hline
		AlexNet \cite{hu2017dee} & 0.7890 \\
		SWP-CNN \cite{hu2017dee} & 0.9310 \\
		RA-CNN \cite{fu2017look} & 0.9250 \\
		MA-CNN \cite{zheng2017recognition}  & 0.9280 \\
		Yaming Wang \cite{wang2018learning} & \textbf{0.9380} \\
		\hline
		DenseNet161 & 0.9292 \\
		DenseNet161-CMP($r$=16) & 0.9371 \\
		VGG16 & 0.9184 \\
		VGG16-CMP($r$=2) & 0.9194 \\
		ResNet152 & 0.9110 \\
		ResNet152\_WoGAP & 0.9249 \\
		ResNet152-CMP($r$=16) & 0.9296 \\
		\hline
	\end{tabular}%
	\label{tab:addlabel}%
\end{table}%

\subsection{Evaluation on the  Cars-196 Dataset}
The Cars-196 dataset contains 16,185 images of 196 car models. This dataset provides ground-truth annotations of bounding boxes, on both training set and test set. In our experiments, we evaluated the proposed method on the Cars-196 dataset with bounding box annotations on 196 car model classes.

1) Experimental Design: We investigate the experimental design of the proposed method, especially the compression factor of the CMP layer.
The compression factor effects the overall performance of the proposed method, and model parameters. We clearly determine these parameters for fine-grained vehicle classification. We set the value of compression factor $r\in\left\{2, 4, 8, 16, 32\right\}$.

Table II represents the comparison of classification results of various DCNNs with different compression factor settings.
For the  ResNet152-CMP and DenseNet161-CMP, We observe that the accuracies are better than corresponding baseline models regardless of the value of compression factor, and the parameters of the model are significantly reduced. Moreover,  especially when the value of compression factor is larger, the performance is better.
For VGG16-CMP, the performance is very sensitive when adjusting the compression factor. Here, we observe that the accuracies generally improve when decreasing the compression factor to 2. However, setting the  compression factor too large (\emph{e.g.}, $r=32$) may hurt the performance.One of the reason is less channel number of compressed feature maps in VGG16-CMP which means less pooled features are generated resulting in losing more information of the corresponding input image when the compression factor is larger.
The highest accuracy of 93.71\% is achieved by the DenseNet161-CMP with setting $r=16$, and the parameter quality is reduced by 64\% with the corresponding baseline.

2) Comparison of Results: Table III shows the comparison of classification results of baseline, state-of-the-art, and proposed models  on the Cars-196 dataset with bounding box annotations.
The VGG16, ResNet152, and DenseNet161 achieve classification accuracy of $91.84\%$, $92.49\%$, and $92.92\%$, respectively.
Moreover, we get some performance gains, \emph{i.e.}, 91.10\% to 92.49\% for ResNet152-WoGAP by remove the GAP layer. This results, which confirm our hypothesis that the GAP may discard some fine-grained information and hurt the performance of the classification.
Furthermore, by applying the CMP layer, all their accuracies are improved considerably, \emph{i.e.}, 92.92\% to 93.71\% for DenseNet161-CMP, 91.84\% to 91.94\% for VGG16-CMP, and 92.49\% to 92.92\% for ResNet152-CMP, the parameters of the model are greatly reduced while improving the results, which shows that the CMP method does improve the performance of the car model classification.
Then, the comparison with state-of-the-art results is also illustrated in Table III. Four recently proposed methods that perform well on this dataset are 93.10\% of the SWP-CNN framework, 92.50\% of RA-CNN framework, 92.80\% of MA-CNN framework, and 93.80\% in [0].
Our DenseNet161-CMP method achieves the accuracy of 93.71\%, which is a very competitive result.

\begin{table*}[htbp]
	\centering
	\caption{COMPARISON OF CLASSIFICATION RESULTS OF VARIOUS DCNNs WITH DIFFERENT PARAMETER SETTINGS ON THE COMPCARS DATASET.}
	\begin{tabular}{c|c|c|c|c|c|c}
		\hline
		compression & \multicolumn{2}{c|}{DenseNet161-CMP} & \multicolumn{2}{c|}{VGG16-CMP} & \multicolumn{2}{c}{ResNet152-CMP} \\
		\cline{2-7}    rate ($r$) & acc.  & params. & acc.  & params. & acc.  & params. \\
		\hline
		2     & 0.9745 & 54.5M & \textbf{0.9780} & 21.4M & 0.9601 & 84.2M \\
		4     & \textbf{0.9789} & 40.6M & 0.9766 & 18.2M & 0.9675 & 72.3M \\
		8     & 0.9776 & 33.6M & 0.9753 & 16.6M & \textbf{0.9701} & 64.8M \\
		16    & 0.9767 & 30.2M & 0.9742 & 15.8M & 0.9649 & 61.6M \\
		32    & 0.9756 & 28.4M & 0.9714 & 15.3M & 0.9625 & 60.0M \\
		\hline
		\hline
		& \multicolumn{2}{c|}{DenseNet161} & \multicolumn{2}{c|}{VGG16} & \multicolumn{2}{c}{ResNet152-WoGAP} \\
		\cline{2-7}    -     & acc.  & params. & acc.  & params. & acc.  & params. \\
		\cline{2-7}          & 0.9715 & 82.3M & 0.9760 & 27.8M & 0.9667 & 109.9M \\
		\hline
	\end{tabular}%
	\label{tab:addlabel}%
\end{table*}%

\begin{table}[htbp]
	\centering
	\caption{COMPARISON OF CLASSIFICATION RESULTS ON THE COMPCARS DATASET WITH BOUNDING BOX ANNOTATIONS.}
	\begin{tabular}{c|c}
		\hline
		Methods & Acc. on Model \\
		\hline
		AlexNet \cite{yang2015large} & 0.8190 \\
		OverFeat \cite{yang2015large} & 0.8790 \\
		GoogLeNet \cite{yang2015large} & 0.9120 \\
		BoxCars \cite{sochor2016boxcars} & 0.8480 \\
		SWP-CNN \cite{hu2017dee} & 0.9760 \\
		\hline
		DenseNet161 & 0.9715 \\
		DenseNet161-CMP($r$=4) & \textbf{0.9789} \\
		VGG16 & 0.9760 \\
		VGG16-CMP($r$=2) & 0.9780 \\
		ResNet152 & 0.9520 \\
		ResNet152\_WoGAP & 0.9667 \\
		ResNet152-CMP($r$=8) & 0.9701 \\
		\hline
	\end{tabular}%
	\label{tab:addlabel}%
\end{table}%

\subsection{Evaluation on the CompCars Dataset}

The CompCars is a hybrid dataset which contains 52,083 images of 431 car models. This dataset also provides ground-truth annotations of bounding boxes, on both the training set and test set. In our experiments, we evaluate the proposed method on the CompCars with bounding box annotations. We train proposed models on classes of 431 car models.

1) Experimental Design: Consistent with settings in evaluation on the Cars-196 Dataset.

Table IV represents the comparison of classification results of various DCNNs with different compression factor settings.
For the DenseNet161-CMP, We observe that the accuracy is better than corresponding baseline models regardless of the value of compression factor, and the parameters of the model have been significantly reduced.
For the VGG16-CMP,  when the compression factor is smaller, the performance will be better than bseline.
For the  ResNet152-CMP, We observe that the accuracy is better than corresponding baseline models when the model parameters were reduced by 46\% between with corresponding baseline.

The highest accuracy of 97.82\% is achieved by the DenseNet161-CMP with setting $r$ = 4, and the model parameters were reduced by 50.7\% between with corresponding baseline.

2) Comparison of Results: Table V shows the comparison of classification  results of baseline, state-of-the-art, and proposed models  on the CompCars dataset with bounding box annotations.

The VGG16, ResNet152, and DenseNet161 achieve classification accuracy of $97.60\%$, $96.67\%$, and $97.15\%$, respectively.Moreover, we get some performance gains, \emph{i.e.}, $95.20\%$ to $96.67\%$ for ResNet152-WoGAP by remove the GAP layer.This results also confirm our hypothesis that the GAP may discard some fine-grained information and hurt the performance of the classification.
Furthermore, by applying the CMP layer, all their accuracies are improved considerably, \emph{i.e.}, $97.15\%$ to $97.82\%$ for DenseNet161-CMP, $97.60\%$ to $97.80\%$ for VGG16-CMP, and $96.67\%$ to $97.01\%$ for ResNet152-CMP, the parameters of the model are greatly reduced while improving the results, which shows that the CMP method does improve the performance of the car model classification.

Then, the comparison with state-of-the-art results is also illustrated on Table IV. Two recently proposed methods that perform well on this dataset are $84.8\%$ of the BoxCars framework and $97.60\%$ of the SWP-CNN framework.
Our DenseNet161-CMP method achieves the best accuracy of $97.89\%$ outperforming all previously-reported results.

\section{Conclusion}
In the paper, we proposed a new layer called as Channel Max Pooling Layer for fine-grained vehicle classification based on neural networks, and analyzed essential difference with max pooling. The proposed layer aims to help neural network learn more general discriminative features to improve the performance of fine-grained vehicle classification. Experimental results on the Stanford Cars-196 dataset and the CompCars dataset confirmed that our method, Channel Max Pooling Layer $\left(1\right)$, can make a network employed CMP Layer obtain better performance than the ones without CMP Layer, $\left(2\right)$ reduce a mass of parameters of neural networks, and $\left(3\right)$ is able to achieve competitive performance with the state-of-the-art performance. Future work includes applying multiple CMP layers in the existing DCNNs, experimenting on different types of networks as well as different kinds of data to evaluate the effectiveness of the CMP layers.







\end{document}